\documentclass{article}
\usepackage{ijcai16}
\usepackage{times}
\usepackage{multirow}
\usepackage{latexsym}
\usepackage{mathrsfs}
\usepackage{graphicx}
\usepackage{subfig}
\usepackage{amsfonts}
\usepackage{amssymb}
\usepackage{amsmath}
\usepackage{bm}
\usepackage{url}
\usepackage{tikz-dependency}
\usepackage{pgfplots}
\usepackage{xcolor}
\usepackage{array}

\allowdisplaybreaks

\newenvironment{itemize*}%
  {\begin{itemize}%
    \setlength{\itemsep}{0pt}%
    \setlength{\parskip}{0pt}}%
  {\end{itemize}}
  \newenvironment{enumerate*}%
  {\begin{enumerate}%
    \setlength{\itemsep}{0pt}%
    \setlength{\parskip}{0pt}}%
  {\end{enumerate}}

\newcommand{\tabincell}[2]{\begin{tabular}{@{}#1@{}}#2\end{tabular}}
\pgfplotsset{compat=1.5}
\newcommand\newcite[1]{\citeauthor{#1} [\citeyear{#1}]}

\def\h{\mathbf{h}}

\def\bx{\mathbf{x}}
\def\by{\mathbf{y}}
\def\cc{\mathbf{c}}
\def\ii{\mathbf{i}}
\def\ff{\mathbf{f}}
\def\oo{\mathbf{o}}

\def\cc{\mathbf{c}}

\def\H{\mathbf{H}}

\def\R{\mathbb{R}}

\title{Modelling Interaction of Sentence Pair with Coupled-LSTMs}

\author{Pengfei Liu \quad Xipeng Qiu\thanks{Corresponding author.} \quad Xuanjing Huang\\
Shanghai Key Laboratory of Intelligent Information Processing, Fudan University\\
School of Computer Science, Fudan University\\
825 Zhangheng Road, Shanghai, China\\
\{pfliu14,xpqiu,xjhuang\}@fudan.edu.cn}

\date{}

\begin{document}
\maketitle
\begin{abstract}
Recently, there is rising interest in modelling the interactions of two sentences with deep neural networks. However, most of the existing methods encode two sequences with separate encoders, in which a sentence is encoded with little or no information from the other sentence. In this paper, we propose a deep architecture to model the strong  interaction of sentence pair with two coupled-LSTMs. Specifically, we introduce two coupled ways to model the interdependences of two LSTMs, coupling the local contextualized interactions of two sentences. We then aggregate these interactions and use a dynamic pooling to select the most informative features. Experiments on two very large datasets demonstrate the efficacy of our proposed architecture and its superiority to state-of-the-art methods.
\end{abstract}

\section{Introduction}

Distributed representations of words or sentences have been widely used in many natural language processing (NLP) tasks, such as text classification \cite{kalchbrenner2014convolutional}, question answering and machine translation \cite{sutskever2014sequence} and so on.
Among these tasks, a common problem is modelling the relevance/similarity of the sentence pair, which is also called text semantic matching.

Recently, deep learning based models is rising a substantial interest in text semantic matching and have achieved some great progresses~\cite{hu2014convolutional,qiu2015convolutional,wan2015deep}.


According to the phases of interaction between two sentences, previous models can be classified into three categories.

\vspace{-1em}
\paragraph{Weak interaction Models}
Some early works focus on sentence level interactions, such as ARC-I\cite{hu2014convolutional}, CNTN\cite{qiu2015convolutional} and so on. These models first encode two sequences with some basic (Neural Bag-of-words, BOW) or advanced (RNN, CNN) components of neural networks separately, and then compute the matching score based on the distributed vectors of two sentences. In this paradigm, two sentences have no interaction until arriving final phase.
\vspace{-1em}
\paragraph{Semi-interaction Models}
Some improved methods focus on utilizing multi-granularity representation (word, phrase and sentence level), such as  MultiGranCNN \cite{yin2015convolutional} and Multi-Perspective CNN \cite{he2015multi}.
Another kind of models use soft attention mechanism to obtain the representation of one sentence by  depending on representation of another sentence, such as  ABCNN \cite{yin2015abcnn}, Attention LSTM\cite{rocktaschel2015reasoning,hermann2015teaching}.
These models can alleviate the weak interaction problem, but are still insufficient to model the contextualized interaction on the word as well as phrase level.
\vspace{-1em}
\paragraph{Strong Interaction Models}
These models directly build an interaction space between two sentences and model the interaction at different positions. ARC-II \cite{hu2014convolutional} and MV-LSTM \cite{wan2015deep}. These models enable the model to easily capture the difference between semantic capacity of two sentences.




 In this paper, we propose a new deep neural network architecture to model the strong interactions of two sentences.
 Different with modelling two sentences with separated LSTMs, we utilize two interdependent LSTMs, called coupled-LSTMs, to fully affect each other at different time steps. The output of coupled-LSTMs at each step depends on both sentences. Specifically, we propose two interdependent ways for the coupled-LSTMs: loosely coupled model (LC-LSTMs) and tightly coupled model (TC-LSTMs). Similar to bidirectional LSTM for single sentence \cite{schuster1997bidirectional,graves2005framewise}, there are four directions can be used in coupled-LSTMs. To utilize all the information of four directions of coupled-LSTMs, we aggregate them and adopt a dynamic pooling strategy to automatically select the most informative interaction signals. Finally, we feed them into a fully connected layer, followed by an output layer to compute the matching score.

 The contributions of this paper can be summarized as follows.
 \begin{enumerate}
   \item Different with the architectures of using similarity matrix, our proposed architecture directly model the strong interactions of two sentences with coupled-LSTMs, which can capture the useful local semantic relevances of two sentences. Our architecture can also capture the multiple granular interactions by several stacked coupled-LSTMs layers.
   \item Compared to the previous works on text matching, we perform extensive empirical studies on two very large datasets. The massive scale of the datasets allows us to train a very deep neural networks. Experiment results demonstrate that our proposed architecture is more effective than state-of-the-art methods.
 \end{enumerate}

\section{Sentence Modelling with LSTM}


Long short-term memory network (LSTM)~\cite{hochreiter1997long} is a type of recurrent neural network (RNN) \cite{Elman:1990}, and
specifically addresses the issue of learning long-term dependencies. LSTM maintains a memory cell that updates and exposes its content only when deemed necessary.

While there are numerous LSTM variants, here we use the LSTM architecture used by \cite{jozefowicz2015empirical}, which is similar to the architecture of \cite{graves2013generating} but without peep-hole connections.

We define the LSTM \emph{units} at each time step $t$ to be a collection of vectors in $\mathbb{R}^d$: an \emph{input gate} $\ii_t$, a \emph{forget gate} $\ff_t$,  an \emph{output gate} $\oo_t$, a \emph{memory cell} $\cc_t$ and a hidden state $\h_t$. $d$ is the number of the LSTM units. The elements of the gating vectors $\ii_t$, $\ff_t$ and $\oo_t$ are in $[0, 1]$.

The LSTM is precisely specified as follows.

\begin{align}
	\begin{bmatrix}
		\mathbf{\tilde{c}}_{t} \\
		\mathbf{o}_{t} \\
		\mathbf{i}_{t} \\
		\mathbf{f}_{t}
	\end{bmatrix}
	&=
	\begin{bmatrix}
		\tanh \\
		\sigma \\
		\sigma \\
		\sigma
	\end{bmatrix}
	T_{\mathbf{A},\mathbf{b}}
	\begin{bmatrix}
		\mathbf{x}_{t} \\
		\mathbf{h}_{t-1}
	\end{bmatrix},\label{eq:lstm1}\\
\mathbf{c}_{t} &=
		\mathbf{\tilde{c}}_{t} \odot \mathbf{i}_{t}
		+ \mathbf{c}_{t-1} \odot \mathbf{f}_{t}, \\
	\mathbf{h}_{t} &= \mathbf{o}_{t}  \odot \tanh\left( \mathbf{c}_{t}  \right)\label{eq:lstm2},
\end{align}
where $\bx_t$ is the input at the current time step;
$T_{\mathbf{A},\mathbf{b}}$ is an affine transformation which depends on parameters of the network $\mathbf{A}$ and $\mathbf{b}$.
$\sigma$ denotes the logistic sigmoid function and $\odot$ denotes elementwise multiplication. Intuitively, the forget gate controls the amount of which each unit of the memory cell is erased, the input gate controls how much each unit is updated, and the output gate controls the exposure of the internal memory state.

The update of each LSTM unit can be written precisely as follows
\begin{align}
(\h_t,\cc_t) &= \mathbf{LSTM}(\h_{t-1},\cc_{t-1},\mathbf{x}_t).
\end{align}
Here, the function $\mathbf{LSTM}(\cdot, \cdot, \cdot)$ is a shorthand for Eq. (\ref{eq:lstm1}-\ref{eq:lstm2}).

\section{Coupled-LSTMs for Strong Sentence Interaction}

To deal with two sentences, one straightforward method is to model them with two separate LSTMs. However, this method is difficult to model local interactions of two sentences. An improved way is to introduce attention mechanism, which has been used in many tasks, such as machine translation~\cite{bahdanau2014neural} and question answering~\cite{hermann2015teaching}.

Inspired by the multi-dimensional recurrent neural network \cite{graves2007multi,graves2009offline,byeon2015scene} and grid LSTM~\cite{kalchbrenner2015grid} in computer vision community, we propose two models to capture the interdependences between two parallel LSTMs, called \textbf{coupled-LSTMs} (C-LSTMs).
\begin{figure}[t]\centering
\subfloat[Parallel LSTMs]{
        \includegraphics[width=0.4\linewidth]{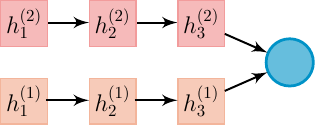}\label{fig:arc-p}
  }\hspace{0.2cm}
  \subfloat[Attention LSTMs]{
        \includegraphics[width=0.4\linewidth]{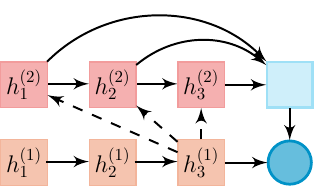}\label{fig:arc-a}
  }\\
  \subfloat[Loosely coupled-LSTMs]{
  \includegraphics[width=0.50\linewidth]{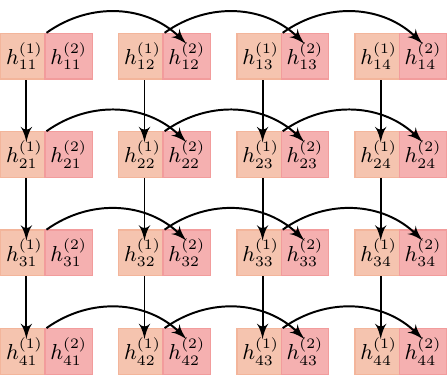}\label{fig:arc-lc}
  }
  \hspace{0.2cm}
  \subfloat[Tightly coupled-LSTMs]{\includegraphics[width=0.41\linewidth]{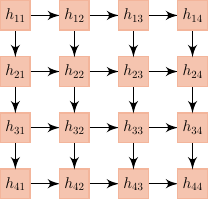}\label{fig:arc-tc}
  }
  \caption{Four different coupled-LSTMs.}\label{fig:arc}
\end{figure}

To facilitate our models, we firstly give some definitions. Given two sequences $X = x_1,x_2,\cdots,x_n$ and $Y = y_1,y_2,\cdots,y_m$, we let $\bx_i \in \R^d$ denote the embedded representation of the word $x_i$. The standard LSTM have one temporal dimension. When dealing with a sentence, LSTM regards the position as time step. At position $i$ of sentence $x_{1:n}$, the output $\h_i$ reflects the meaning of subsequence $x_{0:i}={x_0,\cdots,x_i}$.

To model the interaction of two sentences as early as possible, we define $\h_{i,j}$ to represent the interaction of the subsequences $x_{0:i}$ and $y_{0:j}$.

Figure \ref{fig:arc}(c) and \ref{fig:arc}(d) illustrate our two propose models. For intuitive comparison of weak interaction parallel LSTMs, we also give  parallel LSTMs and attention LSTMs in Figure \ref{fig:arc}(a) and \ref{fig:arc}(b).

We describe our two proposed models as follows.

\subsection{Loosely Coupled-LSTMs (LC-LSTMs)}
To model the local contextual interactions of two sentences, we enable two LSTMs to be interdependent at different positions. Inspired by Grid LSTM~\cite{kalchbrenner2015grid} and word-by-word attention LSTMs \cite{rocktaschel2015reasoning}, we propose a loosely coupling model for two interdependent LSTMs.

More concretely, we refer to $\h_{i,j}^{(1)}$ as the encoding of subsequence $x_{0:i}$ in the first LSTM influenced by the output of the second LSTM on subsequence $y_{0:j}$. Meanwhile, $\h_{i,j}^{(2)}$ is the encoding of subsequence $y_{0:j}$ in the second LSTM influenced by the output of the first LSTM on subsequence $x_{0:i}$

$\h_{i,j}^{(1)}$ and $\h_{i,j}^{(2)}$ are computed as
\begin{align}
\h_{i,j}^{(1)} &= \mathbf{LSTM}^{1}(\mathbf{H}_{i-1}^{(1)},\cc_{i-1,j}^{(1)},\mathbf{\bx}_{i}), \\
\h_{i,j}^{(2)} &= \mathbf{LSTM}^{2}(\mathbf{H}_{j-1}^{(2)},\cc_{i,j-1}^{(2)},\mathbf{y}_{j}),
\end{align}
where
\begin{align}
\mathbf{H}_{i-1}^{(1)} = [\h_{i-1,j}^{(1)},\h_{i-1,j}^{(2)}], \\
\mathbf{H}_{j-1}^{(2)} = [\h_{i,j-1}^{(1)},\h_{i,j-1}^{(2)}].
\end{align}

\subsection{Tightly Coupled-LSTMs (TC-LSTMs)}

The hidden states of LC-LSTMs are the combination of the hidden states of two interdependent LSTMs, whose memory cells are separated. Inspired by the configuration of the multi-dimensional LSTM~\cite{byeon2015scene}, we further conflate both the hidden states and the memory cells of two LSTMs. We assume that $\h_{i,j}$ directly model the interaction of the subsequences $x_{0:i}$ and $y_{0:j}$, which depends on two previous interaction $\h_{i-1,j}$ and $\h_{i,j-1}$, where $i,j$ are the positions in sentence $X$ and $Y$.

We define a tightly coupled-LSTMs \emph{units} as follows.

\begin{align}
	\begin{bmatrix}
		\mathbf{\tilde{c}}_{i,j} \\
		\mathbf{o}_{i,j} \\
		\mathbf{i}_{i,j} \\
		\mathbf{f}_{i,j}^1 \\
		\mathbf{f}_{i,j}^2
	\end{bmatrix}
	&=
	\begin{bmatrix}
		\tanh \\
		\sigma \\
		\sigma \\
		\sigma \\
		\sigma
	\end{bmatrix}
	T_{\mathbf{A},\mathbf{b}}
	\begin{bmatrix}
		\mathbf{x}_{i} \\
        \mathbf{y}_{j} \\
		\mathbf{h}_{i,j - 1} \\
		\mathbf{h}_{i - 1,j}
	\end{bmatrix},\\
	\mathbf{c}_{i,j} &=
		\mathbf{\tilde{c}}_{i,j} \odot \mathbf{i}_{i,j}
		  + [\mathbf{c}_{i,j - 1},\mathbf{c}_{i - 1,j}]^\mathrm{T}
            \begin{bmatrix}
				\mathbf{f}_{i,j}^1 \\
				\mathbf{f}_{i,j}^2 \\
			\end{bmatrix} \\
	\mathbf{h}_{i,j} &= \mathbf{o}_{t}  \odot \tanh\left( \mathbf{c}_{i,j} \right)
\end{align}
where the gating units $\mathbf{i}_{i,j}$ and $\mathbf{o}_{i,j}$
determine which memory units are affected by the inputs through $\mathbf{\tilde{c}}_{i,j}$, and which memory cells are written to the hidden units $\mathbf{h}_{i,j}$.
$T_{\mathbf{A},\mathbf{b}}$ is an affine transformation which depends on parameters of the network $\mathbf{A}$ and $\mathbf{b}$.
In contrast to the standard LSTM defined over time, each memory unit $\cc_{i,j}$ of a tightly coupled-LSTMs has two preceding states
$\mathbf{c}_{i,j-1}$ and $\mathbf{c}_{i-1,j}$ and two corresponding forget gates $\mathbf{f}_{i,j}^1$ and $\mathbf{f}_{i,j}^2$.


\begin{figure*}[t]\centering
  \includegraphics[width=1\linewidth]{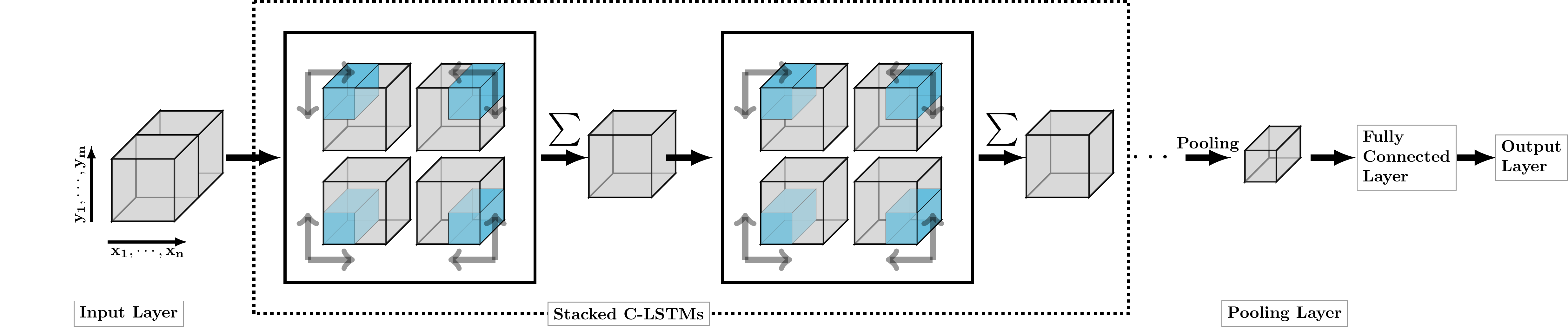}

  \caption{Architecture of coupled-LSTMs for sentence-pair encoding. Inputs are fed to four C-LSTMs followed by an aggregation layer. Blue cuboids represent different contextual information from four directions.}\label{fig:arc-4}
\end{figure*}

\subsection{Analysis of Two Proposed Models}
Our two proposed coupled-LSTMs can be formulated as
\small{
\begin{align}
(\h_{i,j},\cc_{i,j}) &= \textbf{C-LSTMs}(\h_{i-1,j},\h_{i,j-1},\cc_{i-1,j},\cc_{i,j-1},\bx_i,\by_j),
\end{align}
}
where $\textbf{C-LSTMs}$ can be either $\textbf{TC-LSTMs}$ or  $\textbf{LC-LSTMs}$.

 The input consisted of two type of information at step $(i,j)$ in coupled-LSTMs: temporal dimension
$\h_{i-1,j}, \h_{i,j-1}, \cc_{i-1,j}, \cc_{i,j-1}$ and depth dimension
$\mathbf{x}_i,\mathbf{y}_j$. The difference between TC-LSTMs and LC-LSTMs is the dependence of information from temporal and depth dimension.


\paragraph{Interaction Between Temporal Dimensions}
The TC-LSTMs model the interactions at position $(i,j)$ by merging the internal memory $\mathbf{c}_{i-1,j}$ $\mathbf{c}_{i,j-1}$ and hidden state $\mathbf{h}_{i-1,j}$ $\mathbf{h}_{i,j-1}$ along row and column dimensions. In contrast with TC-LSTMs, LC-LSTMs firstly use two standard LSTMs in parallel, producing hidden states $\mathbf{h}^1_{i,j}$ and $\mathbf{h}^2_{i,j}$ along row and column dimensions respectively, which are then merged together flowing next step.

\paragraph{Interaction Between Depth Dimension}
In TC-LSTMs, each hidden state $\mathbf{h}_{i,j}$ at higher layer receives a fusion of information $\mathbf{x}_i$ and $\mathbf{y}_j$, flowed from lower layer. However, in LC-LSTMs, the information $\mathbf{x}_i$ and $\mathbf{y}_j$ are accepted by two corresponding LSTMs at the higher layer separately.

The two architectures have their own characteristics, TC-LSTMs give more strong interactions among different dimensions while LC-LSTMs ensures the two sequences interact closely without being conflated using two separated LSTMs.

\subsubsection{Comparison of LC-LSTMs and word-by-word Attention LSTMs}
The main idea of attention LSTMs is that the representation of sentence X is obtained dynamically based on the alignment degree between the words in sentence X and Y, which is asymmetric unidirectional encoding.
Nevertheless, in LC-LSTM, each hidden state of  each step  is obtained with the consideration of interaction between two sequences with symmetrical encoding fashion.

%
%
%


\section{End-to-End Architecture for Sentence Matching}
In this section, we present an end-to-end deep architecture for matching two sentences, as shown in Figure \ref{fig:arc-4}.

\subsection{Embedding Layer}
To model the sentences with neural model, we firstly need transform the one-hot representation of word into the distributed representation.
All words of two sequences $X = x_1,x_2,\cdots,x_n$ and $Y = y_1,y_2,\cdots,y_m$ will be mapped into low dimensional vector representations, which are taken as input of the network.
%

\subsection{Stacked Coupled-LSTMs Layers}
After the embedding layer, we use our proposed coupled-LSTMs to capture the strong interactions between two sentences. A basic block consists of five layers. We firstly use four directional coupled-LSTMs to model the local interactions with different information flows. And then we sum the outputs of these LSTMs by aggregation layer.
To increase the learning capabilities of the coupled-LSTMs, we stack the basic block on top of each other.

\subsubsection{Four Directional Coupled-LSTMs Layers}

The C-LSTMs is defined along a certain pre-defined direction, we can extend them to access to the surrounding context in all directions.
Similar to bi-directional LSTM, there are four directions in coupled-LSTMs.
{\small\begin{align}
(\h^1_{i,j},\cc^1_{i,j}) &= \textbf{C-LSTMs}(\h_{i-1,j},\h_{i,j-1},\cc_{i-1,j},\cc_{i,j-1},\bx_i,\by_j),\nonumber\\
(\h^2_{i,j},\cc^2_{i,j})  &= \textbf{C-LSTMs}(\h_{i-1,j},\h_{i,j+1},\cc_{i-1,j},\cc_{i,j+1},\bx_i,\by_j),\nonumber\\
(\h^3_{i,j},\cc^3_{i,j})  &= \textbf{C-LSTMs}(\h_{i+1,j},\h_{i,j+1},\cc_{i+1,j},\cc_{i,j+1},\bx_i,\by_j),\nonumber\\
(\h^4_{i,j},\cc^4_{i,j})  &= \textbf{C-LSTMs}(\h_{i+1,j},\h_{i,j-1},\cc_{i+1,j},\cc_{i,j-1},\bx_i,\by_j).\nonumber
\end{align}
}

\subsubsection{Aggregation Layer}
 The aggregation layer sums the outputs of four directional coupled-LSTMs into a vector.

\begin{equation}
\hat{\h}_{i,j} = \sum_{d=1}^{4}\h_{i,j}^{d},
\end{equation}
where the superscript $t$ of $\h_{i,j}$ denotes the different directions.

\subsubsection{Stacking C-LSTMs Blocks}


To increase the capabilities of network of learning multiple granularities of interactions, we stack several blocks (four C-LSTMs layers and one aggregation layer) to form deep architectures.


\subsection{Pooling Layer}

The output of stacked coupled-LSTMs layers is a tensor $\H \in \R^{n\times m \times d}$, where $n$ and $m$ are the lengths of sentences, and $d$ is the number of hidden neurons. We apply dynamic pooling to automatically extract $\R^{p \times q}$ subsampling matrix in each slice $\H_i \in \R^{n \times m}$, similar to \cite{socher2011dynamic}.

More formally, for each slice  matrix $\H_i$, we partition the rows and columns of $\H_i$ into $p \times q$ roughly equal grids. These grid are non-overlapping. Then we select the maximum value within each grid.
Since each slice $\H_i$ consists of the hidden states of one neuron at different positions, the pooling operation can be regarded as the most informative interactions captured by the neuron.

Thus, we get a $p \times q \times d $ tensor, which is further reshaped into a vector.

\subsection{Fully-Connected Layer}

The vector obtained by pooling layer is fed into a full connection layer to obtain a final more abstractive representation.

\subsection{Output Layer}

The output layer depends on the types of the tasks, we choose the corresponding form of output layer.
There are two popular types of text matching tasks in NLP. One is ranking task, such as community question answering. Another is classification task, such as textual entailment.

\begin{enumerate}
  \item For ranking task, the output is a scalar matching score, which is obtained by a linear transformation after the last fully-connected layer.
  \item For classification task, the outputs are the probabilities of the different classes, which is computed by a softmax function after the last fully-connected layer.
\end{enumerate}

\section{Training}

Our proposed architecture can deal with different sentence matching tasks.  The loss functions varies with different tasks.

\paragraph{Max-Margin Loss for Ranking Task}
Given a positive sentence pair $(X,Y)$ and its corresponding negative pair $(X,\hat{Y})$. The matching score $s(X,Y)$ should be larger than $s(X,\hat{Y})$.

For this task, we use the contrastive max-margin criterion \cite{Bordes:2013,socher2013reasoning} to train our models on matching task.

The ranking-based loss is defined as
\begin{equation}
L(X,Y,\hat{Y})=max(0, 1 - s(X,Y) + s(X,\hat{Y})).
\end{equation}
where $s(X,Y)$ is predicted matching score for $(X,Y)$.

\paragraph{Cross-entropy Loss for Classification Task}

 Given a sentence pair $(X,Y)$ and its label $l$. The output $\hat{l}$ of neural network is the probabilities of the different classes. The parameters of the network are trained to minimise the cross-entropy of the predicted and true label distributions.

\begin{equation}
  L(X,Y; \emph{\textbf{l}}, \hat{\emph{\textbf{l}}}) = - \sum_{j=1}^C  \emph{\textbf{l}}_j \log(\hat{\emph{\textbf{l}}}_j),
\end{equation}
where $\emph{\textbf{l}}$ is one-hot representation of the ground-truth label $l$; ${\hat{\emph{\textbf{l}}}}$ is predicted probabilities of labels; $C$ is the class number.

To minimize the objective, we use stochastic gradient descent with the diagonal variant of AdaGrad \cite{duchi2011adaptive}. To prevent exploding gradients, we perform gradient clipping by scaling the gradient when the norm exceeds a threshold \cite{graves2013generating}.

\section{Experiment}
In this section, we investigate the empirical performances of our proposed model on two different text matching tasks: classification task (recognizing textual entailment) and ranking task (matching of question and answer).

\begin{table}[t]  \setlength{\tabcolsep}{3pt}
\centering
\begin{tabular}{|l|*{2}{p{0.18\linewidth}|}}
    \hline
    & MQA & RTE\\\hline
    Embedding size & 100 & 100\\
    Hidden layer size &50 & 50\\
    Initial learning rate& 0.05 & 0.005\\
    Regularization & $5E{-5}$ & $1E{-5}$\\
    Pooling $(p,q)$  & (2,1) & (1,1)\\
    \hline
\end{tabular}
\caption{Hyper-parameters for our model on two tasks.}\label{tab:paramSet}
\end{table}

\subsection{Hyperparameters and Training}

The word embeddings for all of the models are initialized with the 100d GloVe vectors (840B token version, \cite{pennington2014glove}) and fine-tuned during training to improve the performance.
The other parameters are initialized by randomly sampling from uniform distribution in $[-0.1, 0.1]$.

For each task, we take the hyperparameters which achieve the best performance on the development set via an small grid search over combinations of the initial learning rate $[0.05, 0.0005, 0.0001]$, $l_2$ regularization $[0.0, 5E{-5}, 1E{-5}, 1E{-6}]$ and the threshold value of gradient norm [5, 10, 100].
The final hyper-parameters are set as Table \ref{tab:paramSet}.

\subsection{Competitor Methods}
\begin{itemize}
  \item Neural bag-of-words (NBOW): Each sequence as the sum of the embeddings of the words it contains, then they are concatenated and fed to a MLP.
  \item Single LSTM: A single LSTM to encode the two sequences, which is used in \cite{rocktaschel2015reasoning}.
  \item Parallel LSTMs: Two sequences are encoded by two LSTMs separately, then they are concatenated and fed to a MLP.

  \item Attention LSTMs: An attentive LSTM to encode two sentences into a semantic space, which used in \cite{rocktaschel2015reasoning}.
  \item Word-by-word Attention LSTMs: An improvement of attention LSTM by introducing word-by-word attention mechanism, which used in \cite{rocktaschel2015reasoning}.
\end{itemize}

\subsection{Experiment-I: Recognizing Textual Entailment}

Recognizing textual entailment (RTE) is a task to determine the semantic relationship between two sentences. We use the Stanford Natural Language Inference Corpus (SNLI) \cite{bowman-EtAl:2015:EMNLP}. This corpus contains 570K sentence pairs, and all of the sentences and labels stem from human annotators. SNLI is two orders of magnitude larger than all other existing RTE corpora. Therefore, the massive scale of SNLI allows us to train powerful neural networks such as our proposed architecture in this paper.

\begin{table}[!t]\small
\center
\begin{tabular}{|l|*{5}{c|}}
\hline
\textbf{Model} & $k$ & $|\theta|_{M}$ & Train & Test \\ \hline
NBOW & 100 & 80K & 77.9 & 75.1 \\
\tabincell{l}{single LSTM \\\cite{rocktaschel2015reasoning}} & 100 & 111K & 83.7 & 80.9 \\
\tabincell{l}{parallel LSTMs \\\cite{bowman-EtAl:2015:EMNLP}} & 100 & 221K & 84.8 & 77.6 \\

\tabincell{l}{Attention LSTM \\ \cite{rocktaschel2015reasoning}} & 100 & 252K & 83.2 & 82.3 \\
\tabincell{l}{Attention(w-by-w) LSTM\\ \cite{rocktaschel2015reasoning}} & 100 & 252K & 83.7 & 83.5 \\ \hline

LC-LSTMs (Single Direction) & 50 & 45K & 80.8 & 80.5 \\
LC-LSTMs  & 50 & 45K & 81.5 & 80.9 \\
four stacked LC-LSTMs & 50 & 135K & 85.0 & 84.3 \\ \hline

TC-LSTMs (Single Direction) & 50 & 77.5K & 81.4 & 80.1 \\
TC-LSTMs & 50 & 77.5K & 82.2 & 81.6 \\
four stacked TC-LSTMs & 50 & 190K & 86.7 & \textbf{85.1} \\ \hline
\end{tabular}
\caption{Results on SNLI corpus.}\label{tab:res-exp2}
\end{table}

\begin{table*}[!t]
\centering
\begin{tabular}{|c|l|}
\hline
\textbf{Index of Cell} & \textbf{Word or Phrase Pairs} \\
\hline
$3$-th & (in a pool, swimming), (near a fountain, next to the ocean), (street, outside)\\
$9$-th & (doing a skateboard, skateboarding), (sidewalk with, inside), (standing, seated)\\
$17$-th & (blue jacket, blue jacket), (wearing black, wearing white), (green uniform, red uniform)\\
$25$-th & (a man, two other men), (a man, two girls), (an old woman, two people)\\ \hline
\end{tabular}
\caption{Multiple interpretable neurons and the word-pairs/phrase-pairs captured by these neurons.} \label{tab:exp-pair}
\end{table*}

\subsubsection{Results}
Table \ref{tab:res-exp2} shows the evaluation results on SNLI.
The $3$rd column of the table gives the number of parameters of different models without the word embeddings.

Our proposed two C-LSTMs models with four stacked blocks outperform all the competitor models, which indicates that our thinner and deeper network does work effectively.

Besides, we can see both LC-LSTMs and TC-LSTMs benefit from multi-directional layer, while the latter obtains more gains than the former. We attribute this discrepancy between two models to their different mechanisms of controlling the information flow from depth dimension.

Compared with attention LSTMs, our two models achieve comparable results to them using much fewer parameters (nearly $1/5$).  By stacking C-LSTMs, the performance of them are improved significantly, and the four stacked TC-LSTMs achieve $85.1\%$ accuracy on this dataset.

Moreover, we can see TC-LSTMs achieve better performance than LC-LSTMs on this task, which need fine-grained reasoning over pairs of words as well as phrases.

\begin{figure}[t]\centering

  \subfloat[3rd neuron]{\includegraphics[height=5cm]{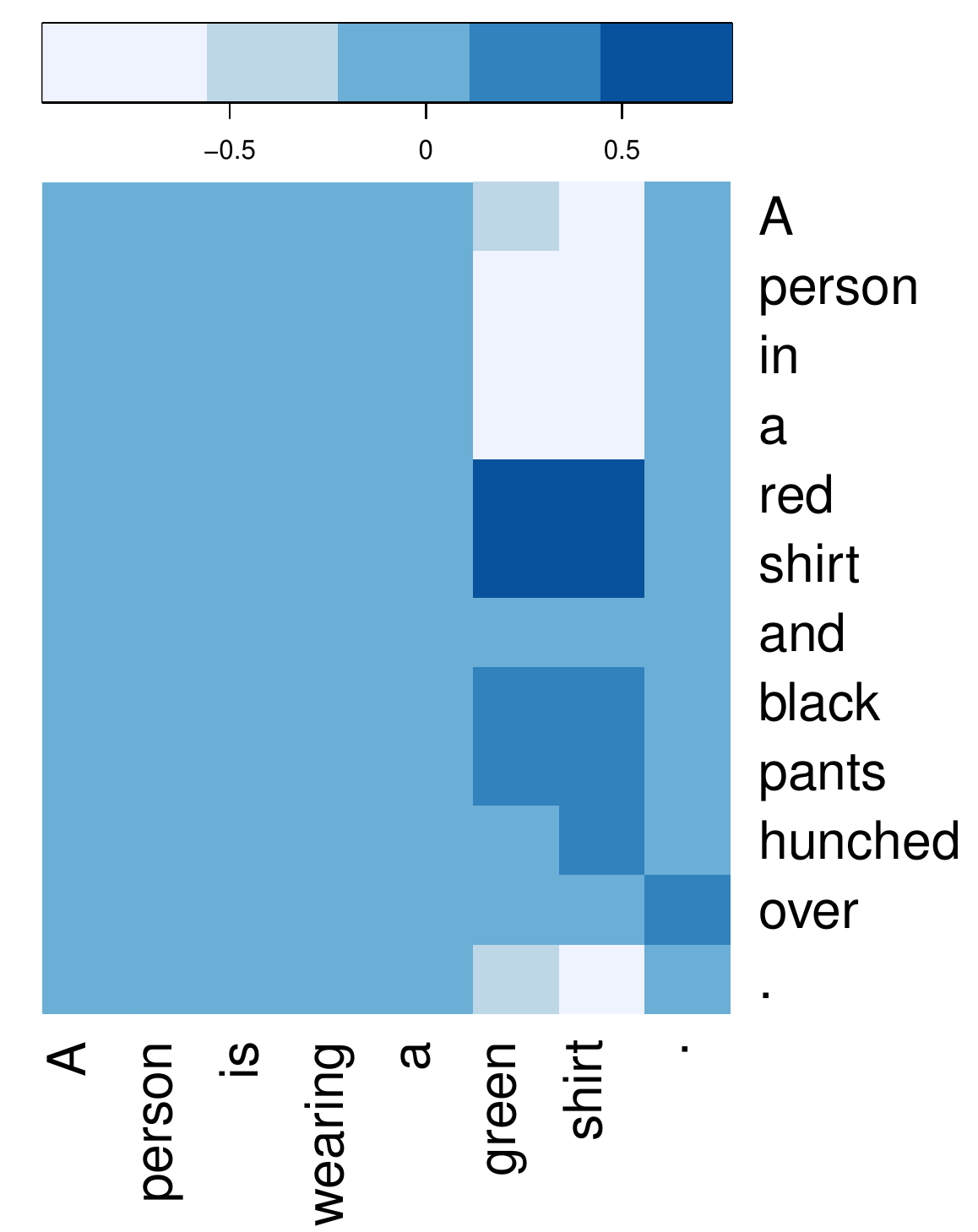}
  }\hspace{0.3cm}
  \subfloat[17th neuron]{
  \includegraphics[height=5cm]{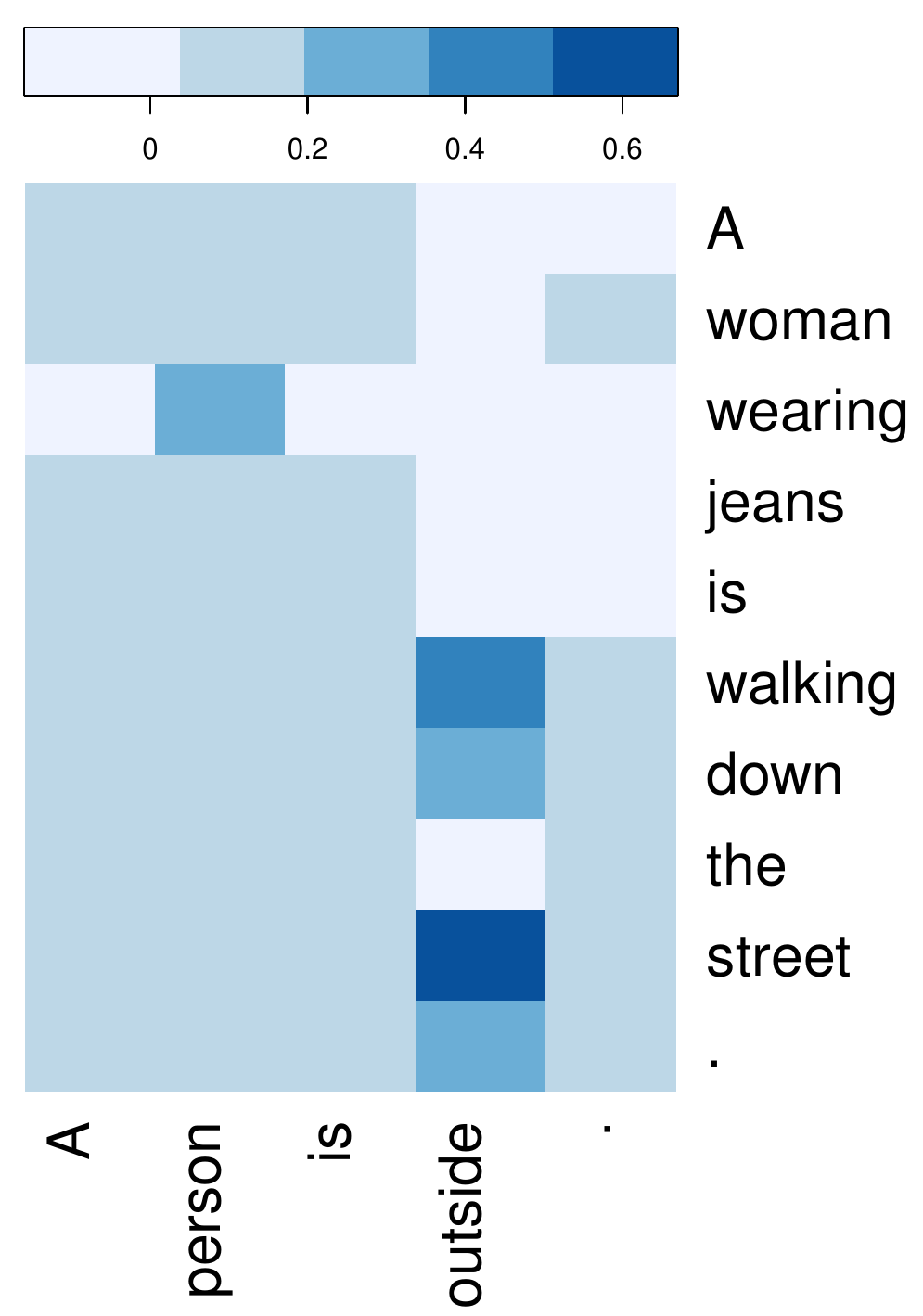}
  }

  \caption{Illustration of two interpretable neurons and some word-pairs capture by these neurons. The darker patches denote the corresponding activations are higher.}\label{fig:case-studt}
\end{figure}

\subsubsection{Understanding Behaviors of Neurons in C-LSTMs}

To get an intuitive understanding of how the C-LSTMs work on this problem, we examined the neuron activations in the last aggregation layer while evaluating the test set using TC-LSTMs. We find that some cells are bound to certain roles.





Let $h_{i,j,k}$ denotes the activation of the $k$-th neuron at the position of $(i,j)$, where $i \in \{1,\ldots,n\}$ and $j \in \{1,\ldots,m\}$. By visualizing the hidden state $\h_{i,j,k}$ and analyzing the maximum activation, we can find that there exist multiple interpretable neurons.
For example, when some contextualized local perspectives are semantically related at point $({i},{j})$ of the sentence pair, the activation value of hidden neuron $h_{{i},{j},k}$ tend to be maximum, meaning that the model could capture some reasoning patterns.

Figure \ref{fig:case-studt} illustrates this phenomenon. In Figure \ref{fig:case-studt}(a), a neuron shows its ability to monitor the local contextual interactions about color.
The activation in the patch, including the word pair  ``\texttt{(red, green)}'', is much higher than others. This is informative pattern for the relation prediction of these two sentences, whose ground truth is contradiction. An interesting thing is there are two words describing color in the sentence `` \texttt{A person in a red shirt and black pants hunched over.}''. Our model ignores the useless word ``\texttt{black}'', which indicates that this neuron selectively captures pattern by contextual understanding, not just word level interaction.

In Figure \ref{fig:case-studt}(b), another neuron shows that it can capture the local contextual interactions, such as ``\texttt{(walking down the street, outside)}''. These patterns can be easily captured by pooling layer and provide a strong support for the final prediction.

Table \ref{tab:exp-pair} illustrates multiple interpretable neurons and some representative word or phrase pairs which can activate these neurons. 
These cases show that our models can capture contextual interactions beyond word level.

\subsubsection{Error Analysis}
Although our models C-LSTMs are more sensitive to the discrepancy of the semantic capacity between two sentences, some semantic mistakes at the phrasal level still exist.
For example, our models failed to capture the key informative pattern when predicting the entailment sentence pair ``\texttt{A girl \textbf{takes off her shoes} and eats blue cotton candy/The girl is eating while \textbf{barefoot}.}''

Besides, despite the large size of the training corpus, it's still very different to solve some cases, which depend on the combination of the world knowledge and context-sensitive inferences.
For example, given an entailment pair ``\texttt{a man grabs his crotch during a political demonstration/The man is making a crude gesture}'', all models predict ``\texttt{neutral}''.
This analysis suggests that some architectural improvements or external world knowledge are necessary to eliminate all errors instead of simply scaling up the basic model.

\subsection{Experiment-II: Matching Question and Answer}
Matching question answering (MQA) is a typical task for semantic matching.
Given a question, we need select a correct answer from some candidate answers.

In this paper, we use the dataset collected  from Yahoo! Answers with the getByCategory function provided in Yahoo! Answers API, which produces $963,072$ questions and corresponding best answers. We then select the pairs in which the length of questions and answers are both in the interval $[4,30]$, thus obtaining $220,000$ question answer pairs to form the positive pairs.

For negative pairs, we first use each question's best answer as a query to retrieval top $1,000$ results from the whole answer set with Lucene, where $4$ or $9$ answers will be selected randomly to construct the negative pairs.

The whole dataset is divided into training, validation and testing data with proportion $20:1:1$. Moreover, we give two test settings: selecting the best answer from 5 and 10 candidates respectively.

\begin{table}[!t]\small
\center
\begin{tabular}{|l|*{4}{c|}}
\hline
\textbf{Model} & $k$  & P@1(5) & P@1(10) \\ \hline
Random Guess & - & 20.0 & 10.0 \\
NBOW & 50 &  63.9 & 47.6 \\
single LSTM  & 50 &  68.2 & 53.9 \\
parallel LSTMs & 50 &  66.9 & 52.1 \\
Attention LSTMs & 50 &  73.5 & 62.0 \\\hline
LC-LSTMs (Single Direction) & 50 &  75.4 & 63.0 \\
LC-LSTMs & 50 &  76.1 & 64.1 \\
three stacked LC-LSTMs & 50 &  \textbf{78.5} & \textbf{66.2} \\ \hline

TC-LSTMs (Single Direction) & 50 &  74.3 & 62.4 \\
TC-LSTMs & 50 &  74.9 & 62.9 \\
three stacked TC-LSTMs & 50 &  77.0 & 65.3 \\ \hline
\end{tabular}
\caption{Results on Yahoo question-answer pairs dataset.}\label{tab:res-exp1}
\end{table}

\subsubsection{Results}
Results of MQA are shown in the Table \ref{tab:res-exp1}. For our models,  due to stacking block more than three layers can not make significant improvements on this task, we just use three stacked C-LSTMs.

By analyzing the evaluation results of question-answer matching in table \ref{tab:res-exp1}, we can see
strong interaction models (attention LSTMs, our C-LSTMs) consistently outperform the weak interaction models (NBOW, parallel LSTMs) with a large margin, which suggests the importance of modelling strong interaction of two sentences.

Our proposed two C-LSTMs surpass the competitor methods and C-LSTMs augmented with multi-directions layers and multiple stacked blocks fully utilize multiple levels of abstraction to directly boost the performance.

Additionally, LC-LSTMs is superior to TC-LSTMs. The reason may be that MQA is a relative simple task, which requires less reasoning abilities, compared with RTE task. Moreover, the parameters of LC-LSTMs are less than TC-LSTMs, which ensures the former can avoid suffering from overfitting on a relatively smaller corpus.


\section{Related Work}
Our architecture for sentence pair encoding can be regarded as strong interaction models, which have been explored in previous models.

An intuitive paradigm is to compute similarities between all the words or phrases of the two sentences. \newcite{socher2011dynamic} firstly used this paradigm for paraphrase detection. The representations of words or phrases are learned  based on recursive autoencoders. \newcite{wan2015deep} used LSTM to enhance the positional contextual interactions of the words or phrases between two sentences. The input of LSTM for one sentence does not involve another sentence.

A major limitation of this paradigm is the interaction of two sentence is captured by a pre-defined similarity measure. Thus, it is not easy to increase the depth of the network. Compared with this paradigm, we can stack our C-LSTMs to model multiple-granularity interactions of two sentences.


\newcite{rocktaschel2015reasoning} used two LSTMs equipped with attention mechanism to capture the iteration between two sentences. This architecture is asymmetrical for two sentences, where the obtained final representation is sensitive to the two sentences' order.

Compared with the attentive LSTM, our proposed C-LSTMs are symmetrical and model the local contextual interaction of two sequences directly.

\section{Conclusion and Future Work}
In this paper, we propose an end-to-end deep architecture to capture the strong interaction information of sentence pair. Experiments on two large scale text matching tasks demonstrate the efficacy of our proposed model and its superiority to competitor models.
Besides, our visualization analysis revealed that multiple interpretable neurons in our proposed models can capture the contextual interactions of the words or phrases.

In future work, we would like to incorporate some gating strategies into the depth dimension of our proposed models, like highway or residual network, to enhance the interactions between depth and other dimensions thus training more deep and powerful neural networks.

\bibliographystyle{named}
\bibliography{../nlp,../ours}

\end{document}